# ParsTranslit: Truly Versatile Tajik-Farsi Transliteration

**Rayyan Merchant**
Zoorna Institute
Miami, USA
`rayyan.merchant@gmail.com`

**Kevin Tang**
Heinrich-Heine Universität / Düsseldorf, Germany
University of Florida / Gainesville, USA
`kevin.tang@hhu.de`

## Abstract

As a digraphic language, the Persian language utilizes two written standards: Perso-Arabic in Afghanistan and Iran, and Tajik-Cyrillic in Tajikistan. Despite the significant similarity between the dialects of each country, script differences prevent simple one-to-one mapping, hindering written communication and interaction between Tajikistan and its Persian-speaking "siblings". To overcome this, previously-published efforts have investigated machine transliteration models to convert between the two scripts. Unfortunately, most efforts did not use datasets other than those they created, limiting these models to certain domains of text such as archaic poetry or word lists. A truly usable transliteration system must be capable of handling varied domains, meaning that suck models lack the versatility required for real-world usage. The contrast in domain between data also obscures the task's true difficulty. We present a new state-of-the-art sequence-to-sequence model for Tajik-Farsi transliteration trained across all available datasets, and present two datasets of our own. Our results across domains provide clearer understanding of the task, and set comprehensive comparable leading benchmarks. Overall, our model achieves chrF++ and Normalized CER scores of 87.91 and 0.05 from Farsi to Tajik and 92.28 and 0.04 from Tajik to Farsi. Our model, data, and code are available at `https://github.com/merchantrayyan/ParsTranslit/`.

## 1 Introduction

As a digraphic language, the Persian language utilizes two written standards: Perso-Arabic in Afghanistan and Iran (henceforth Farsi[1]), and Tajik-Cyrillic in Tajikistan (henceforth Tajik). Tajik-Farsi transliteration is the task of converting between these two incongruent written standards, made possible by the enduring high mutual intelligibility between standard Persian varieties (Merchant et al., 2025). However, despite the similarities between the two varieties, few in Tajikistan are able to access written content in Farsi, which dwarfs Tajik in Internet presence (SadraeiJavaheri et al., 2024). As such, development of an accurate Tajik-Farsi transliteration system has the potential to 1) connect Tajikistan with the rest of the Persophone world and to 2) allow for the application of Farsi NLP tools to Tajik text. Compared to Tajik, these Farsi NLP tools are much more developed, with tools such as Parsivar (Mohtaj et al., 2018) and DadmaToolsV2 (Jafari et al., 2025) capable of part-of-speech tagging, lemmatization, sentiment analysis, dependency parsing, and more.

Our paper presents the following contributions:

1. We introduce a state-of-the-art Tajik-Farsi transliteration model, the first to be trained across a wide variety of contexts. As a compact and versatile model, it is capable of transliterating both modern-day news articles and centuries-old poetic works.

2. We are the first to evaluate our model's performance in comparison with others across multiple datasets and contexts, setting benchmarks for future work. In doing so, we also reveal what types of data models find challenging to transliterate between Tajik and Farsi.

3. We introduce two new datasets of our own creation for this task. The first is a novel parallel entity name dataset for Tajik-Farsi derived from the ParaNames dataset (Sälevä and Lignos, 2024). The second is another version of Rumi's Masnavi, an important piece of Persian literature, manually checked for major discrepancies. Both are publicly accessible to support further work in this area.

4. We propose the insertion of contextual marker characters as a data augmentation technique

---
[1] For the purposes of this paper and brevity's sake, Farsi will be used as an umbrella term for Afghan and Iranian Persian.

for this task, as used in Arabic-Arabizi transliteration (Shazal et al., 2020). Our experiments revealed that doing so yielded noticeable improvements over unmodified data.

The paper is organized as follows. In Section 2, we review the challenges of Tajik-Farsi transliteration, such as orthographic idiosyncrasies. Section 3 provides an overview of previous Tajik–Farsi transliteration systems and Section 4 details each of the available datasets used in this study. In Sections 5 and 6, we present our experiment setup and evaluation methods, respectively. Finally, Section 7 presents and discusses the overall results, and Section 8 concludes the paper and outlines directions for future work. The code and data are publicly available at `https://github.com/merchantrayyan/ParsTranslit/`.

## 2 Transliteration Challenges

Several orthographic idiosyncrasies complicate this task, as detailed in Merchant et al. (2025) and Megerdoomian and Parvaz (2008). The main issues are summarized below.

**Unwritten and Ambiguous Vowels** As a derivation of the Arabic abjad, Perso-Arabic omits most short vowels, making one-to-one mapping impossible. Farsi has retained several "redundant" characters from Arabic whose pronunciations are now identical (Perry, 2005). In Tajik orthography, these historical distinctions are not preserved, so words that are written differently in Farsi often appear identical in Tajik, producing homographs from what are heterographs in Farsi.

**"Ezafe" and Phrasal Boundaries** The *Ezafe* is a Persian linking morpheme realized as an и ('i') in Tajik and typically-unwritten diacritic in Farsi. This morpheme ties noun phrases together, attaching to adjectives and denoting possession. As a result, similar to Farsi grapheme-to-phoneme systems, a Farsi→Tajik transliteration system must be capable of detecting this morphological feature (Rahmati and Sameti, 2024).

**Transliteration and Capitalization of Named Entities** Unlike the Tajik-Cyrillic script, the Perso-Arabic script lacks capitalization. Farsi→Tajik transliteration therefore requires the use of named entity recognition to correctly capitalize named entities.

**Zero-Width Non-Joiner and Word Boundaries**
The Zero-Width Non-Joiner (ZWNJ) is an invisible Farsi character used to visibly separate affixes from the word to which they are affixed. Common examples include the plural marker ها *'ha'* and direct object marker را *'ra'* (Megerdoomian and Parvaz, 2008). When the ZWNJ is used, these affixes appear separate, but remain part of the same token. In informal texts, this character is often substituted with a space or removed entirely, complicating word-level tokenization (Merchant et al., 2025). This phenomenon is not observed in Tajik, where affixes are simply attached to the preceding word.

## 3 Previous Work

Table 1 provides an overview of previous Tajik–Farsi transliteration systems. The top half of the table lists the datasets each system has incorporated, ranging from word lists and literary corpora to news and blogs. The bottom half summarizes system properties, including transliteration direction, availability of digraphic training and test data, and whether code or models were released. Each of the systems is described in more detail in the following subsections.

| Datasets | Previous Systems | | | | | | Ours |
|---|---|---|---|---|---|---|---|
| | 3.1 | 3.2 | 3.3 | 3.4 | 3.5 | 3.6 | 5 |
| Word List | | | ✓ | | | | |
| Dictionary | | | | ✓ | ✓ | | ✓ |
| Shahnameh | | | | | | ✓ | ✓ |
| Masnavi (either version) | | | | ✓ | ✓ | | ✓ |
| Assorted Poetry | | | | ✓ | ✓ | | ✓ |
| BBC News | | | | ✓ | ✓ | | ✓ |
| Blogs | | | | | ✓ | | ✓ |
| ParaNames | | | | | | | ✓ |
| Direction: Tajik→Farsi | ✓ | ✓ | ✓ | ✓ | ✓ | ✓ | ✓ |
| Direction: Farsi→Tajik | ✗ | ✗ | ✓ | ✗ | ✓ | ✓ | ✓ |
| Digraphic Training Data | ✗ | ✗ | ✓ | ✓ | ✓ | ✓ | ✓ |
| Digraphic Test Data | ✗ | ✗ | ✗ | ✓ | ✓ | ✓ | ✓ |
| Model/Data Availability | ✗ | ✗ | ✗ | ✓ | ✓ | ✓ | ✓ |

Table 1: Overview of previous systems and our system in terms of a breakdown of the training datasets (top), and the system properties (bottom) – transliteration direction, data type and model/data availability

### 3.1 Tajik→Farsi Finite-State Transducer System

Megerdoomian and Parvaz (2008) developed a Tajik→Farsi transliteration system utilizing a finite-state transducer (FST). The FST first generates a number of possible Farsi transliterations for the input Tajik sequence. A combination of morphological analysis and a dictionary lookup are then used to determine a match. If no exact match is found, the system references letter frequencies to select a most likely alternative. When evaluated on a test corpus made up of news articles from Radio Ozodi, the Tajik branch of Radio Free Europe, this system generated on average 6.27 alternative spellings for each token and achieved 89.8% accuracy in transliterating a document from Tajik to Farsi.

### 3.2 A Patented Tajik→Farsi System

A patented mathematical-based Tajik→Farsi transliteration system was developed by the Academy of Sciences of the Republic of Tajikistan (Usmanov et al., 2008; Graschenko, 2008; Graschenko and Fomin, 2008; Graschenko, 2009; Graschenko et al., 2009). Only limited information about the inner workings of the system can be found on their website[2] and the accompanying papers. To the best of the authors' knowledge, this remains the only transliteration system developed by researchers in Tajikistan and has not been cited in prior English-language papers as of September 2025.

### 3.3 Bidirectional Statistical Transliteration Model

Davis (2012) trained a statistical model on a 3,503 digraphic word list and utilized character-level language models built on a Tajik online news source and the Farsi Bijankhan corpus (Amiri et al., 2007). Evaluation was conducted through two tasks: part of speech (POS) tagging and machine translation. A Tajik POS tagger trained on model-transliterated text achieved an accuracy of 92.52%. In the machine translation task, the model first transliterated text from Tajik to Farsi, and then the transliterated texts was translated into English with Google Translate; the system achieved a BLEU score of 0.2349 (Papineni et al., 2001).

### 3.4 Tajik→Farsi Transformer System from Russia

Seredkina (2024) proposed a Tajik→Farsi transliteration system trained on parallel texts available in both scripts, including poetry, news articles, and a Tajik-Farsi dictionary. They evaluated both LSTM and transformer models on this dataset, reporting Edit Distance ratios of 0.990 and 0.989, respectively. The implementation, data, and pretrained models are publicly accessible on GitHub (stibiumghost, 2022).

### 3.5 Bidirectional Grapheme-to-Phoneme Model

Merchant et al. (2025) trained a Grapheme-to-Phoneme (G2P) transformer model using the DeepPhonemizer (Schäfer et al., 2023) implementation of work by Yolchuyeva et al. (2019). They used the dataset by Seredkina (2024) and supplemented this with their own corpus, ParsText, (Merchant and Tang, 2024), which consists of blogs and news articles. As shown in Table 1, this system was the first to report direct evaluation metrics in both directions. The model achieved chrF++ scores of 58.70 (Farsi→Tajik) and 64.43 (Tajik→Farsi), with sequence accuracies of 33.99% and 34.37%, respectively. Excluding the Zero-Width Non-Joiner (ZWNJ) markedly improved Tajik→Farsi performance to 74.20 chrF++ and 50.46% accuracy. The code and pretrained models are publicly available on GitHub (merchantrayyan, 2025).

### 3.6 Bidirectional Transformer System trained on the *Shahnameh*

SadraeiJavaheri et al. (2024) trained character-level GRU and transformer models exclusively on an aligned version of the Shahnameh ("The Book of Kings"), a medieval Persian epic by Ferdowsi. They evaluated these models against OpenAI's `gpt-3.5-turbo` using a three-shot prompting setup. The transformer model outperformed both the GRU and ChatGPT, achieving average edit distances of 0.88 (Tajik→Farsi) and 1.05 (Farsi→Tajik). The corpus, code, and trained models are publicly available on GitHub.[3]

As Table 1 shows, previous systems differ widely in both dataset coverage and methodological transparency. While some models were trained on curated word lists or individual literary sources, few have combined diverse corpora. Moreover, bidirectional transliteration remains limited, with only a subset of systems making code and data publicly available. These gaps motivate our own system, described in Section 5, which integrates multiple text sources, supports both directions, and provides open resources for future research.

---

[2]`https://tajpers.narod.ru/index_e.htm`

[3]`https://github.com/language-ml/Tajiki-Shahname`

## 4 Datasets

As seen above, previous efforts typically only utilized datasets they had created themselves, limiting their models to certain domains of text like poetry or word lists. By utilizing all the datasets currently available as of September 2025, of both others' creation as well as our own, we collected datasets spanning diverse domains such as: blogs, news articles, poetry, word lists, and named entities. Table 2 provides a general overview of each of the available datasets including the number of digraphic pairs, and the average number of tokens and characters for each script. These are described in further detail below.

| Domain / Datasets | | # of Pairs | Farsi | | Tajik | |
|---|---|---|---|---|---|---|
| | | | Avg. # Tokens | Avg. # Char. | Avg. # Tokens | Avg. # Char. |
| Poetry | Shahnameh | 68,206 | 5.68 | 24.77 | 5.47 | 29.25 |
| | Masnavi | 39,011 | 6.12 | 25.81 | 5.75 | 31.43 |
| | Assorted Poetry | 156,576 | 6.31 | 27.20 | 6.01 | 33.10 |
| Prose | Dr Blog | 1,554 | 12.33 | 62.40 | 11.84 | 72.02 |
| | Jamujam Blog | 819 | 15.79 | 82.21 | 15.34 | 94.34 |
| | Assorted Prose | 23,992 | 19.56 | 102.45 | 18.62 | 118.95 |
| Names | Places | 11,179 | 1.48 | 9.46 | 1.45 | 10.46 |
| | Organizations | 4,185 | 1.39 | 9.05 | 1.31 | 9.72 |
| | People | 23,988 | 1.77 | 9.96 | 1.76 | 10.93 |
| **Dictionary** | | 49,758 | 1.38 | 6.70 | 1.00 | 7.63 |

Table 2: Overview of the datasets used as training data. Text domain in bold are described in Section 6.2.

### 4.1 Parallel Texts (Seredkina, 2024)

As stated earlier in Section 3.4, this dataset was sourced from texts published in both scripts, and provides additional author details for each text. This proved quite useful when determining the amount of overlap between it and other available datasets. We describe the three sub-datasets they below.

**Assorted Poetry** A collection of poetry by notable historical Persian poets was compiled. While Seredkina (2024) claims that works by contemporary Tajikistani poets Bozor Sobir and Farzona are included, we found that none of the entries in the collection are attributed to them.

**Assorted Prose** An assorted set of prose was compiled from articles and web pages from several websites such as BBC News that published versions in each script.

**Dictionary** A word list was compiled from the contents of the Explanatory Dictionary of the Tajik Language (Yunusova, 1969). Each entry includes a Persian transliteration for each entry, including Russian loanwords.

### 4.2 ParsText (Merchant and Tang, 2024)

Similar to Seredkina (2024), Merchant and Tang (2024) sourced data from online webpages for their publicly-available corpus, ParsText.

**Personal Blogs** Texts were extracted from two personal blogs[4] written by native Persian speakers who could write in both scripts and wrote on diverse topics.

**BBC News Articles** 23 Tajik news articles that had been published in both scripts were extracted from the British Broadcasting Corporation (BBC) website. Upon comparison with the dataset from Seredkina (2024), it was found that all of these articles were present in the Assorted Prose dataset (Section 4.1). Due to this overlap, we excluded this subset of ParsText from our final dataset entirely.

### 4.3 Shahnameh ("The Book of Kings")

SadraeiJavaheri et al. (2024) provide an publicly-available and aligned version of Ferdowsi's Shahnameh, a 10th–11th century epic poem often considered the pinnacle of Persian literature. As no dual-script versions were readily available online, they used optical character recognition on a printed Tajik version and aligned it with an online Farsi version[5] using an alignment approach similar to Seredkina (2024).

### 4.4 Two New Datasets

We collected two datasets of our own from previously-unexplored sources: Wikipedia entity names and the Masnavi written by Rumi. Both datasets are available on Github at https://anonymous.4open.science/r/ParsTranslit-FB30.

**Tajik-Farsi subset of ParaNames** ParaNames[6], a multilingual name resource (MIT License) built from Wikipedia entity records (Sälevä and Lignos, 2024), was used to extract entries with both Tajik and Farsi names. We then applied Seredkina (2024)'s code, which filters pairs based on one-to-one mappings between unambiguous consonants, yielding the first Tajik–Farsi parallel name resource

---
[4]https://dariussthoughtland.blogspot.com/, https://jaamjam.blogspot.com/
[5]https://ganjoor.net/ferdousi/shahname
[6]https://github.com/bltlab/paranames

– 39,352 pairs across three entity types (person, location, organization). To the authors' knowledge, this is the first named entity dataset created for Tajik-Farsi transliteration.

**Masnavi**   We extracted Rumi's Masnavi, a Tajik poetic work, from the website of the Center of Information Technologies and Telecommunication in Tajikistan,[7] and aligned it with a Farsi version from Ganjoor[8]. Passages were aligned by number, and all line discrepancies were manually corrected by an L2 Tajik-speaking author. Although Seredkina (2024) also provides a version of Masnavi, we exclude it due to unclear provenance and encourage researchers to choose either version at their discretion.

## 5 Experiment Setup

### 5.1 Data Preprocessing

To normalize the different datasets (Sec. 4), all punctuation and any characters not belonging to the Perso-Arabic or Tajik-Cyrillic scripts were removed, and Tajik text was lowercased in both directions. However, inconsistently written Perso-Arabic diacritics and ZWNJ from the Farsi text were kept during training, and were only removed during evaluation. Preliminary experiments showed that models trained with diacritics performed noticeably better on the same test set than those trained without, indicating that diacritics, though inconsistently written, provide useful signals rather than noise. Similar gains were observed for the Tajik-Cyrillic hyphen character, which was also removed during evaluation and is often (though not always) used to join two words, such as в-аз 'v-az' ('and from') from ва 'va' ('and') and аз 'az' ('from') (Perry, 2005).

### 5.2 Tokenization and Contextual Markers

The text was tokenized at the character level, using a contextual token "_" to mark spaces between words and "@" to mark word boundaries, as positional cues have been shown to improve character mappings (Megerdoomian and Parvaz, 2008). Following Shazal et al. (2020), who used a similar approach for Arabic transliteration, we observed comparable gains. Before evaluation, the model outputs were detokenized and all contextual markers removed.

---

[7] https://termcom.tj/
[8] https://ganjoor.net/moulavi/masnavi

### 5.3 Model Training

Similar to Seredkina (2024), Merchant et al. (2025), and SadraeiJavaheri et al. (2024), we trained a Seq2Seq transformer model but differed in that we opted to use the Fairseq framework[9] (Ott et al., 2019). Prior to training, the data were binarized and vocabularies built using `fairseq-preprocess`. Each dataset was split into training, development, and test sets (80%, 10%, 10%). Training was conducted with 10-fold cross-validation for 20 epochs on a 2024 MacBook Pro (M3, 16 GB) in both directions (Tajik→Farsi, Farsi→Tajik), totaling around ~100 GPU hours.

### 5.4 Hyperparameters

We used similar hyperparameters to those described by SadraeiJavaheri et al. (2024) and Merchant et al. (2025), with two small differences. First, we set our batch size to 128, as Wu et al. (2021) demonstrated that a batch size of at least 128 allows transformer models to achieve stronger performance on character-level tasks. Second, we conducted very small-scale tests to determine an optimal learning rate, settling on 0.0007 as shown in the Appendix (Table 7). Training beyond 20 epochs yielded only marginal gains. See Appendix (Table 6) for an overview of final hyperparameters.

## 6 Evaluation Method

### 6.1 Model Comparisons

In the following sections, we will refer to our model as 'ParsTranslit'. Despite the number of previous Tajik-Farsi transliteration systems, we were unable to compare our model with several of them due to their unavailability and incompatible metrics. Megerdoomian and Parvaz (2008) and Graschenko et al. (2009) relied on checking accuracy of individual word tokens, while Davis (2012) evaluated downstream tasks. As the datasets they used are unavailable, we are also unable to calculate comparable results using our own model.

As a result, we restrict our direct comparison to models for which both datasets and evaluation metrics are available, namely: the DeepPhonemizer model (Merchant et al., 2025), the model of SadraeiJavaheri et al. (2024), and the unidirectional TG2FA system of Seredkina (2024). We note, however, that the training data for both DeepPhonemizer and TG2FA partially overlap with our test set. This overlap is likely to lead to artificially inflated performance for these baselines across all shared datasets (see Table 1).

---

[9] https://github.com/facebookresearch/fairseq

Furthermore, due to challenges in replicating the implementation of SadraeiJavaheri et al. (2024), we report their results only as presented in the original paper. Specifically, we compare our system against their reported numbers on the Shahnameh dataset, where their train–development–test split is closely aligned with our own. This ensures that the results remain directly comparable despite the lack of a fully reproduced evaluation pipeline.

## 6.2 Grouping Datasets by Domain

As different text domains have been shown to pose distinct challenges for NLP tasks even within standard English (Liu et al., 2024), similar effects may be expected in our datasets, which span multiple domains and centuries. The datasets in Section 4 were therefore grouped into four domains to evaluate model performance on each type. They are as follows: *Dictionary* (Section 4.1), *Prose* (Assorted Prose (Section 4.1); Blog Posts (Section 4.2), *Poetry* (Assorted Poetry (Section 4.1); Masnavi (Section 4.4); Shahnameh (Section 4.3)), and *Names* (People; Places; Organizations) (Section 4.4).

## 6.3 Metrics

We use the following metrics to measure the quality of model transliteration: chrF(++), Character Error Rate (CER), and Sequence Accuracy, as these are commonly-used metrics for transliteration and have been reported previously by Merchant et al. (2025) and SadraeiJavaheri et al. (2024). Since a single incorrect character does not pose a challenge to intelligibility, chrF and CER allow us to evaluate how intelligible transliterations are, while Sequence Accuracy serves as a harsher sequence-level metric. This is particularly important for our proper name and dictionary subsets which consist of less than two word tokens per pair (See Table 2).

### 6.3.1 chrF and chrF++

*Character F-score (chrF)* measures the F-score of character n-grams by comparing overlaps between the output and reference (Popović, 2015). As a character-level metric, it is language-independent. *chrF++* extends chrF by incorporating both character and word unigrams and bigrams. Popović (2017) showed that chrF++ correlates more closely with direct human assessment of machine translation quality than chrF.

### 6.3.2 Character Error Rate (CER) and Normalized CER

*CER* measures the average edit distance between each output and reference, with lower values indicating fewer edits required. *Normalized CER* extends this by dividing the edit distance by sequence length, accounting for sequences ranging from single tokens to entire sentences.

### 6.3.3 Sequence accuracy with and without whitespace

*Sequence accuracy* (Acc%) is the percentage of model outputs that exactly match their references. We also report sequence accuracy with all whitespace removed in both output and reference (Acc% (No WS)), as doing so resulted in noticeable increases in performance (See Section 7.4).

## 7 Results and Discussions

### 7.1 Model Comparison across domains

We present model and baseline performance across domains in Tables 3 and 4, with performance across individual datasets available in Appendix (see Tables 8 and 9). We report on the mean results from the 10-fold split.

**Overall, our ParsTranslit model outperforms the DP in both directions and the TG2FA model going from Tajik to Farsi.**

A sole exception appears in prose domain, where the TG2FA model outperforms our model when transliterating from Tajik to Farsi. The ~4% difference in chrF(++) and Sequence Accuracy is likely caused due to training on the Assorted Prose dataset, which makes up the majority of the prose domain. However, we note that this does not occur for other shared datasets, suggesting the prose domain somehow benefited more from the overlap than other domains.

### 7.2 Model Comparison on the Shahnameh

We compare the ParsTranslit, DP, and TG2FA models against the character error rate (CER) results[10] from SadraeiJavaheri et al. (2024) evaluated on the Shahnameh dataset, as summarized in Table 5. ParsTranslit performs comparably to their model. In the Farsi→Tajik direction, ParsTranslit achieves a CER of 0.93, outperforming the SadraeiJavaheri et al. (2024)'s model (1.05). In contrast, in the Tajik→Farsi direction, the SadraeiJavaheri et al.

---
[10]Reported as Mean Edit Distance in SadraeiJavaheri et al. (2024), though the metric is equivalent to CER.

| Subset | chrF | | chrF++ | | CER | | Normalized CER | | Acc% | | Acc% (No WS) | |
|---|---|---|---|---|---|---|---|---|---|---|---|---|
| | DP | ParsTranslit | DP | ParsTranslit | DP | ParsTranslit | DP | ParsTranslit | DP | ParsTranslit | DP | ParsTranslit |
| Poetry | 65.24 | **92.52** | 58.14 | **90.36** | 4.30 | **0.90** | 0.13 | **0.03** | 1.84 | **53.74** | 3.22 | **58.52** |
| Prose | 64.40 | **86.56** | 56.44 | **83.14** | 13.47 | **7.55** | 0.13 | **0.06** | 4.88 | **18.25** | 5.40 | **20.91** |
| Dictionary | 66.27 | **89.33** | 60.16 | **87.55** | 1.56 | **0.36** | 0.20 | **0.05** | 24.70 | **76.91** | 36.34 | **77.00** |
| Names | 27.63 | **71.78** | 21.81 | **66.83** | 4.29 | **1.50** | 0.40 | **0.15** | 4.05 | **40.82** | 4.14 | **41.44** |
| **Overall** | 63.83 | **90.34** | 56.57 | **87.91** | 4.57 | **1.36** | 0.17 | **0.05** | 5.28 | **52.98** | 7.81 | **56.57** |

Table 3: Farsi→Tajik model performance across different data subsets and across metrics. The score in bold indicates the best performing model for each metric and subset.

| Subset | chrF | | | chrF++ | | | CER | | | Normalized CER | | | Acc% | | | Acc% (No WS) | | |
|---|---|---|---|---|---|---|---|---|---|---|---|---|---|---|---|---|---|---|
| | DP | TG2FA | ParsTranslit | DP | TG2FA | ParsTranslit | DP | TG2FA | ParsTranslit | DP | TG2FA | ParsTranslit | DP | TG2FA | ParsTranslit | DP | TG2FA | ParsTranslit |
| Poetry | 84.73 | 93.21 | **95.63** | 78.69 | 91.19 | **93.98** | 2.16 | 0.86 | **0.60** | 0.08 | 0.03 | **0.02** | 15.64 | 55.75 | **66.35** | 33.22 | 65.56 | **76.50** |
| Prose | 86.25 | **95.59** | 91.64 | 80.54 | **94.16** | 89.85 | 5.97 | **2.09** | 6.39 | 0.07 | **0.03** | 0.04 | 11.11 | **41.96** | 38.25 | 19.17 | **51.21** | 47.11 |
| Dict. | **94.14** | 85.14 | 91.22 | **86.86** | 78.82 | 85.73 | 0.54 | 0.55 | **0.38** | 0.08 | 0.08 | **0.06** | 59.13 | 62.75 | **72.52** | **86.82** | 74.41 | 82.69 |
| Names | 37.41 | 44.57 | **80.08** | 31.05 | 38.28 | **75.82** | 3.40 | 2.75 | **1.02** | 0.34 | 0.29 | **0.11** | 7.43 | 13.17 | **53.68** | 9.48 | 14.15 | **54.48** |
| **Overall** | 83.72 | 92.02 | **94.00** | 77.80 | 90.08 | **92.28** | 2.34 | 1.1 | **1.01** | 0.11 | 0.06 | **0.04** | 20.18 | 51.30 | **63.90** | 36.82 | 60.41 | **72.99** |

Table 4: Tajik→Farsi model performance across different data subsets and across metrics. The score in bold indicates the best performing model for each metric and subset.

(2024)'s model performs similarly to ParsTranslit (0.88 vs. 0.90). The DP and TG2FA models perform considerably worse: DP yields 1.56 CER (Farsi→Tajik) and 2.55 CER (Tajik→Farsi), while TG2FA reaches 1.09 CER (Tajik→Farsi).

| Model | CER | |
|---|---|---|
| | Farsi to Tajik | Tajik to Farsi |
| ParsTranslit | **0.93** | 0.90 |
| SadraeiJavaheri et al. (2024) | 1.05 | **0.88** |
| DP | 1.56 | 2.55 |
| TG2FA | N/A | 1.09 |

Table 5: Model performance on Shahnameh dataset in both directions. The score in bold indicates the best performing model for each direction.

Our model's stronger performance in the Farsi→Tajik direction suggests that this direction benefits from the larger amount of training data available, possibly because it requires predicting a greater number of additional characters. In contrast, in the Tajik→Farsi direction, where both models perform similarly, the results indicate that the advantage of additional data may plateau.

### 7.3 Which domains are the most difficult for all models?

We find that **entity names were the greatest challenge for all models**, exhibiting sharp declines in across all metrics with Normalized CER in both directions dropping ~0.10 and ~0.20 compared to the overall score for ParsTranslit and DP, respectively (see Tables 3 and 4). Notably, the DP and TG2FA models performed substantially worse compared to our ParsTranslit model, suggesting that Tajik-Farsi transliteration models solely trained on parallel texts fail to generalize well to entity names. This may have further impacted model performance on prose given the large amount of foreign named entities in BBC news articles (see Section 4.1). As this is the first time entity names have been directly evaluated in this paradigm, future efforts should continue investigating entity names separately from full texts.

When looking solely at ParsTranslit's unbiased performance, **prose appears more difficult than poetry in both directions**. In Tables 3, and 4, model accuracy steeply drops from 53.75% to 18.25% (Farsi→Tajik) and from 66.35% to 38.25% (Tajik→Farsi) between poetry and prose, despite only minor differences chrF(++) scores between the two domains. This could potentially be reflective of the breadth of topics and named entities prose covers compared to poetry, but could also be an effect of the much larger sequence sizes.

### 7.4 Which direction is harder to transliterate?

As Tajik sequences contain many more characters than Farsi sequences on average (see Table 2), directly comparing transliteration directions using character-level metrics such as chrF and CER may prove misleading, even when accounting for sequence length with Normalized CER. To avoid this, we evaluated character-independent Sequence Accuracy, and discovered that **both the DP and ParsTranslit models found transliterating from Tajik to Farsi to be easier than from Farsi to Tajik**. Just by switching direction from Farsi→Tajik to Tajik→Farsi, the DP model's Sequence Accuracy drastically increased ~46% and the ParsTranslit model improved by ~10%. This aligns with previous findings by Merchant et al. (2025) and SadraeiJavaheri et al. (2024).

Interestingly, calculating accuracy after removing whitespaces from the output and references (Acc% (No WS)) resulted in increases for all models in both directions. However, each direction responded differently, with meager gains (overall less than ~4%) for Farsi→Tajik and substantial gains (overall at least ~10%) for Tajik→Farsi. In this direction, the DP model benefited the most from whitespace removal, experiencing increases of around ~15% compared to the ~10% seen by ParsTranslit and TG2FA. The greater degree to which Tajik→Farsi accuracy increased after whitespace removal compared to Farsi→Tajik suggests that **models seem to find word boundaries in Farsi harder to predict than in Tajik**. This can potentially be attributed to alternative contracted forms and inconsistent inclusion of the invisible Zero-Width Non-Joiner character across the different datasets.

## 8 Conclusion

This paper advances Tajik–Farsi transliteration along several dimensions. We presented the first comprehensive overview of available datasets for Tajik–Farsi transliteration. We introduced a state-of-the-art model capable of handling both contemporary news and historical poetic texts, and we benchmarked its performance against prior work across diverse datasets and contexts. Our approach achieves state-of-the-art performance, with overall chrF++ and normalized CER scores of 87.91/0.05 (Farsi→Tajik) and 92.28/0.04 (Tajik→Farsi), across multiple domains of text. Our results set new baselines for the field, while also revealing domains that remain particularly challenging, underscoring the need for domain-aware evaluation and training.

In addition, we released two new resources: a parallel entity name dataset derived from ParaNames and a carefully revised version of Rumi's Masnavi. Together with our evaluation framework, these datasets provide a foundation for more systematic future work. We also proposed the use of contextual marker characters for data augmentation, showing measurable gains over unmodified training data.

Beyond achieving state-of-the-art chrF++ and CER scores, our analysis highlights limitations of standard translation and transliteration metrics, since they do not fully capture task-specific challenges such as equally valid contracted forms, the detection of Ezafe, and named entity consistency. Addressing these gaps will require not only additional data but also new evaluation protocols tailored to the linguistic particularities of Tajik–Farsi transliteration, enabling more reliable progress in Tajik–Farsi transliteration research. To support such extensions, we make our model, code and data available.

## Limitations

**Non-native Persian Speakers:** The authors are not native speakers of Persian, and so may not be aware of all crucial orthographic and linguistic features that must be taken into account. Nevertheless, we believe our work goes beyond previous work in its scope. We plan to involve native speakers of both Tajik and Farsi in future work for dataset curation and analysis.

**Model Size and Hardware:** We intentionally focus on training a smaller model as this provides faster inference and more versatile application potential. However, we also lack the hardware necessary to investigate more intensive model hyperparameters, which could yield important insights.

**Preprocessing Tools:** Like previous efforts, our work focuses solely on our model and has not evaluated the impact of including Ezafe detection (Doostmohammadi et al., 2020; Jafari et al., 2025) and Named Entity detection tools (Jalali Farahani and Ghassem-Sani, 2021; Jafari et al., 2025) in our preprocessing tasks on model performance. As our metrics indirectly demonstrate ParsTranslit's ability to correctly predict the Ezafe in many cases, we believe direct evaluation of Ezafe with manually-curated data to be the next step.

## Ethical Considerations

The involved university does not require IRB approval for this kind of study, which uses publicly available data without involving human participants. We do not see any other concrete risks concerning dual use of our research results. Of course, in the long run, any research results on AI methods could potentially be used in contexts of harmful and unsafe applications of AI. But this danger is rather low in our concrete case.

## CRediT authorship contribution statement

We follow the CRediT taxonomy[11]. Conceptualization: [RM, KT]; Data curation: [RM]; Formal Analysis: [RM, KT]; Investigation: [RM, KT]; Methodology: [RM, KT]; Supervision: [KT]; Visualization: [RM]; and Writing – original draft: [RM, KT] and Writing – review & editing: [RM, KT].

---

[11]https://credit.niso.org/

## A Appendix

### A.1 Hyperparameters

| Hyperparameter | Value |
| --- | --- |
| Batch Size | 128 |
| Learning Rate | 0.0007 |
| Dropout | 0.1 |
| Layers | 2 |
| Heads | 4 |
| Embedding Dimension | 256 |
| Epochs | 20 |
| Learning Rate Scheduler | Reduce on Plateau |
| Optimizer | Adam |

Table 6: Training hyperparameters

| **Learning Rate** | **Loss** |
| --- | --- |
| 7e-4 | 0.12 |
| 6e-4 | 0.12 |
| 4e-4 | 0.12 |
| 3e-4 | 0.121 |
| 2e-4 | 0.134 |

Table 7: Validation loss across different learning rates during hyperparameter experiments using a Tajik→Farsi model. All other hyperparameters were kept as reported.

### A.2 Model performance on individual datasets

| Subset | chrF | | chrF++ | | CER | | Normalized CER | | Acc% | | Acc% (NoWS) | |
|---|---|---|---|---|---|---|---|---|---|---|---|---|
| | DP | ParsTranslit | DP | ParsTranslit | DP | ParsTranslit | DP | ParsTranslit | DP | ParsTranslit | DP | ParsTranslit |
| **Poetry: Shahnameh** | 65.61 | **92.76** | 57.80 | **90.73** | 4.25 | **0.93** | 0.14 | **0.03** | 1.78 | **59.22** | 3.37 | **63.76** |
| **Poetry: Masnavi** | 65.20 | **91.29** | 58.53 | **89.22** | 4.13 | **0.96** | 0.14 | **0.03** | 2.37 | **51.74** | 4.19 | **55.06** |
| **Poetry: Assorted Poetry** | 65.11 | **92.71** | 58.18 | **90.48** | 4.35 | **0.88** | 0.13 | **0.03** | 1.73 | **51.85** | 2.91 | **57.10** |
| **Prose: Dr Blog** | 58.68 | **85.56** | 51.54 | **82.37** | 11.09 | **5.16** | 0.17 | **0.07** | 5.11 | **29.92** | 5.37 | **32.48** |
| **Prose: Jamujam Blog** | 64.58 | **84.56** | 57.80 | **81.27** | 15.09 | **8.88** | 0.14 | **0.08** | 4.89 | **13.18** | 5.14 | **16.12** |
| **Prose: Assorted Prose** | 64.62 | **86.67** | 56.59 | **83.23** | 13.57 | **7.66** | 0.13 | **0.06** | 4.87 | **17.66** | 5.41 | **20.32** |
| **Names: Places** | 38.32 | **73.75** | 30.69 | **69.43** | 3.51 | **1.51** | 0.36 | **0.17** | 8.98 | **39.00** | 9.11 | **39.89** |
| **Names: Organizations** | 28.37 | **69.40** | 21.92 | **65.13** | 4.27 | **1.76** | 0.47 | **0.18** | 1.76 | **47.64** | 1.85 | **48.31** |
| **Names: People** | 22.75 | **71.23** | 17.87 | **66.13** | 4.66 | **1.46** | 0.41 | **0.14** | 2.19 | **40.46** | 2.25 | **40.95** |
| **Dictionary: Dictionary** | 66.27 | **89.33** | 60.16 | **87.55** | 1.56 | **0.36** | 0.20 | **0.05** | 24.70 | **76.91** | 36.34 | **77.00** |
| **Overall** | 63.83 | **90.34** | 56.57 | **87.91** | 4.57 | **1.36** | 0.17 | **0.05** | 5.28 | **52.98** | 7.81 | **56.57** |

Table 8: Farsi→Tajik model performance on each dataset. The score in bold indicates the best performing model for each metric and dataset.

| Subset | chrF | | | chrF++ | | | CER | | | Normalized CER | | | Acc% | | | Acc% (NoWS) | | |
|---|---|---|---|---|---|---|---|---|---|---|---|---|---|---|---|---|---|---|
| | DP | TG2FA | ParsTranslit | DP | TG2FA | ParsTranslit | DP | TG2FA | ParsTranslit | DP | TG2FA | ParsTranslit | DP | TG2FA | ParsTranslit | DP | TG2FA | ParsTranslit |
| **Poetry: Shahnameh** | 81.61 | 90.92 | **92.48** | 74.61 | 88.76 | **90.53** | 2.55 | 1.09 | **0.90** | 0.10 | 0.04 | **0.03** | 11.58 | 51.35 | **57.15** | 27.80 | 58.50 | **65.00** |
| **Poetry: Masnavi** | 84.30 | 92.79 | **96.30** | 78.40 | 90.94 | **95.16** | 2.16 | 0.86 | **0.43** | 0.08 | 0.03 | **0.02** | 16.52 | 55.92 | **73.60** | 34.44 | 64.74 | **80.49** |
| **Poetry: Assorted Poetry** | 86.04 | 94.20 | **96.71** | 80.32 | 92.18 | **95.04** | 1.99 | 0.76 | **0.51** | 0.07 | 0.03 | **0.02** | 17.18 | 57.62 | **68.55** | 35.28 | 68.85 | **80.52** |
| **Prose: Dr Blog** | 83.70 | **91.48** | 91.26 | 77.71 | **90.22** | 89.89 | 5.39 | **3.02** | 3.34 | 0.09 | **0.05** | 0.05 | 15.72 | 47.32 | **49.10** | 25.32 | 52.29 | **55.23** |
| **Prose: Jamujam Blog** | 84.90 | **91.65** | 89.80 | 79.60 | **89.63** | 87.57 | 7.82 | **5.20** | 6.90 | 0.09 | **0.05** | 0.06 | 10.03 | **29.80** | 24.90 | 16.84 | **37.85** | 32.72 |
| **Prose: Assorted Prose** | 86.39 | **95.86** | 91.71 | 80.68 | **94.46** | 89.92 | 5.95 | **1.92** | 6.57 | 0.07 | **0.02** | 0.04 | 10.85 | **42.03** | 37.99 | 18.85 | **51.60** | 47.07 |
| **Names: Places** | 50.84 | 56.84 | **81.45** | 43.74 | 50.71 | **77.75** | 2.69 | 2.32 | **1.08** | 0.30 | 0.26 | **0.14** | 13.10 | 20.88 | **52.47** | 17.18 | 23.04 | **53.94** |
| **Names: Organizations** | 37.13 | 43.20 | **74.83** | 31.80 | 37.76 | **70.74** | 3.52 | 3.12 | **1.44** | 0.41 | 0.37 | **0.16** | 5.96 | 10.58 | **55.25** | 6.86 | 11.15 | **56.34** |
| **Names: People** | 31.28 | 39.14 | **80.19** | 25.34 | 32.93 | **75.74** | 3.70 | 2.89 | **0.91** | 0.35 | 0.28 | **0.10** | 5.09 | 10.09 | **53.95** | 6.41 | 10.60 | **54.40** |
| **Dictionary: Dictionary** | **94.14** | 85.14 | 91.22 | **86.86** | 78.82 | 85.73 | 0.54 | 0.55 | **0.38** | 0.08 | 0.08 | **0.06** | 59.13 | 62.75 | **72.52** | **86.82** | 74.41 | 82.69 |
| **Overall** | 83.72 | 92.02 | **94.00** | 77.80 | 90.08 | **92.28** | 2.34 | 1.10 | **1.01** | 0.11 | 0.06 | **0.04** | 20.18 | 51.30 | **63.90** | 36.82 | 60.41 | **72.99** |

Table 9: Tajik→Farsi model performance on each dataset. The score in bold indicates the best performing model for each metric and dataset.